# Cross-domain Transfer of Defect Features in Technical Domains Based on Partial Target Data

Tobias Schlagenhauf[1], Tim Scheurenbrand[1]

[1] *Karlsruhe Institute of Technology, Karlsruhe, 76131, Germany*
*tobias.schlagenhauf@kit.edu*
*tim.scheurenbrand@kit.edu*

**ABSTRACT**

A common challenge in real-world classification scenarios with sequentially appending target domain data is insufficient training datasets during the training phase. Therefore, conventional deep learning and transfer learning classifiers are not applicable especially when individual classes are not represented or are severely underrepresented at the outset. Domain Generalization approaches reach their limits when domain shifts become too large, making them occasionally unsuitable as well. In many (technical) domains, however, it is only the defect/ worn/ reject classes that are insufficiently represented, while the non-defect class is often available from the beginning. The proposed classification approach addresses such conditions. Following a contrastive learning approach, a CNN encoder is trained with a modified triplet loss function using two datasets: Besides the non-defective target domain class (= 1st dataset), a state-of-the-art labeled source domain dataset that contains highly related classes (e.g., a related manufacturing error or wear defect) but originates from a (highly) different domain (e.g., different product, material, or appearance) (= 2nd dataset) is utilized. The approach learns the classification features from the source domain dataset while at the same time learning the differences between the source and the target domain in a single training step, aiming to transfer the relevant features to the target domain. The classifier becomes sensitive to the classification features and – by architecture – robust against the highly domain-specific context. The approach is benchmarked in a technical and a non-technical domain and shows convincing classification results. In particular, it is shown that the domain generalization capabilities and classification results are improved by the proposed architecture, allowing for larger domain shifts between source and target domains.

## 1. INTRODUCTION

The detection of defective parts is of great interest in technical domains. Deep Learning classifiers have become an important tool e.g., for the evaluation of manufactured products or the wear monitoring of machine components (Hamadache et al., 2019). In production scenarios there is often an imbalance of defective and non-defective parts, e.g., worn-out ball screw drive (BSD) spindles are rare compared to faultless BSD spindles and reject parts are rare compared to good parts. Related to the lifetime of a component (or production line) this effect is further exacerbated as shown in Figure 1. The figure depicts a schematic lifetime progression of a component subject to wear (Schlagenhauf et al., 2022) and the availability of samples of the non-defective class (green) and the defective class (red). While non-defective parts are available from 0 % lifetime on, the defective class is only available after the initial occurrence of (visually) detectable wear. This is a common situation in industrial scenarios, for processes where no data have been recorded so far. The main issue and motivation for this paper is that to train a model to classify e.g. defects, the training of the classifier must have been completed prior to the initial occurrence of e.g., wear[1]. Figure 2 depicts the corresponding phases of model training, inference, and improvement of an exemplary classification model. Conventional Deep Learning classifiers require an individual training dataset, containing all classes (i.e., non-defective and defective classes) (Ben-David et al., 2010). The same is true for Transfer-Learning approaches, even though there exists a reduced need for data from the target domain. Consequently, these approaches cannot be utilized for a novel domain/ component with zero data from a specific class at the time of model training. Approaches that manage entirely without target domain data are known as Domain Generalization approaches. Solely trained in the source domain, they are subsequently used in the target domain without further training (Zhou et al., 2021). However, their ability for cross-domain generalization requires the domains to be highly similar regarding their features as well as their context (Ben-David et al., 2010; Torralba & Efros, 2011; Wang et al., 2021). This similarity requirement cannot always be guaranteed in technical domains. Even when the image elements of interest, e.g. defects are similar across domains, the context may vary strongly. Furthermore, these approaches do not exploit all the data already available during training, since they are only trained with source-domain data, although data of the *non-defective* target domain class would already be available in most cases. If the domain shift between source and target domains is too large, so-called Domain Adaptation methods may be used instead. These Transfer Learning approaches use two datasets for training: Firstly, the model is pretrained using an already available source domain dataset. Secondly, a (small) target domain dataset is used to adapt the model to the target domain. These approaches are particularly data efficient regarding the target domain since often a small target domain dataset is sufficient if the model can generalize well enough across the domains (Jaiswal et al., 2020; Pan & Yang, 2010; Shen et al., 2022; Zhang et al., 2022). However, like the conventional classifiers, these approaches rely on target domain data from *all classes*, so they can only be trained *after the initial occurrence* of visually detectable wear too (Figure 1).

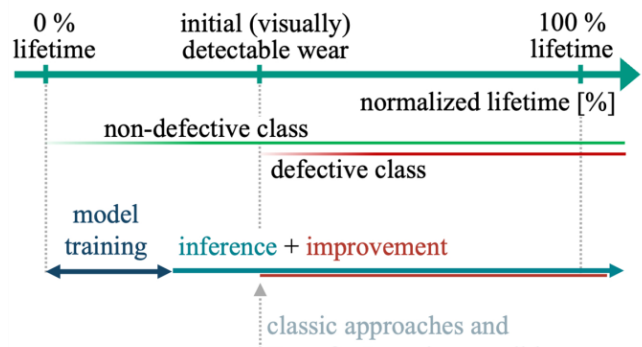

Figure 1: Timeline of model training, improvement, and inference in relation to a wear subjected component's normalized lifetime

We assume that both the source and target domains contain the same classes (closed set). Additionally, the features of the defect to be classified are assumed to only differ marginally between the source domain and the target domain, while the domains themselves may have very different appearances and origins. This allows for a greater difference (domain gap) between source and target domains while at the same time keeping the cost of target domain dataset creation low. Nevertheless, this is a domain generalization approach since the model learns entirely without target domain images of the defective class.

The use-case depicted in Figure 2 motivates our approach: The top row shows a selection of images from the non-defective class of the Severstal strip steel defect dataset (Severstal, 2020). The second row depicts four images from the defective class of the same dataset, showing surface disruptions. These two classes form the source domain dataset, which was already given in the state-of-the-art. The third row shows four images of a BSD spindle in good condition without visible wear (Schlagenhauf, 2021). The last row depicts the same part at a later point in time, now showing surface disruptions after the occurrence of visually detectable wear (known as *pitting*). The task is to classify a random image of the BSD spindle surface into defective and non-defective (the two classes are highlighted by the green border). Since data of defective BSDs are not available during training (see Figure 1), the model is trained using the Severstal dataset (= source domain, row 1 & 2) in addition to the non-defective BSD class (= target domain, row 3). The training data is indicated by the blue outline in Figure 2.

In line with our assumptions, the classification feature of the defective *pitting* class shares domain-independent similarities across the two domains. The appearance of the raw strip steel surface, on the other hand, differs from the heavily machined surface of the ball screw spindle, resulting in a domain shift from the source domain to the target domain. The observation of domain-independent defective class features and domain-specific features of the components themselves (in a non-defective state) is also made by (Rombach et al., 2022).

To summarize the two main characteristics of the here presented approach are that no data from the defective class is available in the target domain and nonetheless *we use the*

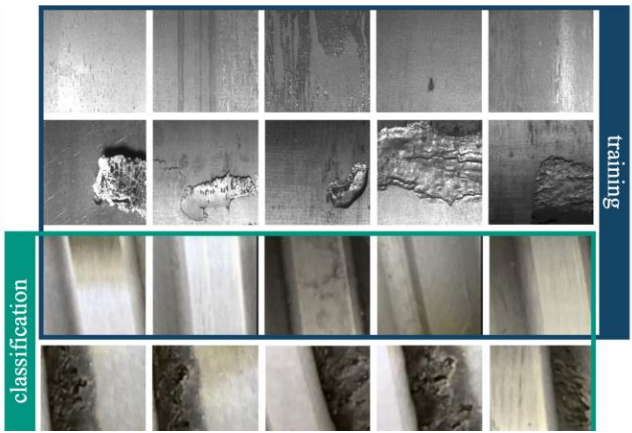

Figure 2: (non-)defective samples from a strip steel defect dataset (row 1 & 2) and BSD wear dataset (row 3 & 4).

---

[1] The example can also be applied to the initial occurrence of a reject part in a production line. Unless a reject has been produced, a classifier cannot be trained with data of it.

*non-defective data of the target domain* together with the data (defective and non-defective) from the source domain to bridge the domain gap. The second characteristic is that we explicitly *alter the formulation of the classical triplet loss function* to bridge the domain gap.

Besides technical domains, the here presented observation holds for other domains as well: In the food industry for the detection of bruises on fruits and vegetables (Siddiqui, 2015) or diseased leaves (Chohan et al., 2020). In the healthcare sector for the detection of disease types (Maqsood et al., 2019). Or in the context of product remanufacturing (e.g., batteries) during the initial diagnosis of products (Kaiser et al., 2021).

## 2. RELATED WORK

### 2.1. Object Detection

Modern object detectors achieve striking performance when train and test data are sampled from the same or similar distributions (Zhang et al., 2022). However, if source and target domains differ, the generalization abilities of state-of-the-art detectors lack (Y. He et al., 2019; Zhang et al., 2021, 2022). One main reason described in the literature is the fact, that an object's background (= context) is related to the object itself (Zhang et al., 2022). Recent studies investigate domain generalization in object detection (DGOD), where detectors trained in source domains are evaluated on unknown target domains (Zhang et al., 2022). To the best of our knowledge, there are no approaches that use partial data from one target domain class, to improve the generalization capabilities of knowledge previously learned from source domain data.

### 2.2. Domain Generalization and Domain Adaptation

Domain Generalization (DG) considers scenarios where target data is inaccessible during model training (Zhou et al., 2021). DG was initially introduced in a medical background where conventional classifiers (trained using data from historic patients) were not able to generalize to new patients due to a distribution shift between different patients' data (Blanchard et al., 2011). For a formal definition of DG, let $X$ be the input (feature) space (e.g., images) and $Y$ the corresponding target (label) space of a domain defined as joint distribution $P_{XY}$ on $X \times Y$. DG intends to learn a domain-invariant prediction model $f: X \rightarrow Y$ using only source domain data $S$ such that the prediction error on an unseen target domain $T$ is minimized. (Wang et al., 2021; Zhou et al., 2021) The generalization abilities of DG approaches are strongly dependent on the similarity of source and target datasets $S, T$ as well as their distributions (Ben-David et al., 2010; Torralba & Efros, 2011). State-of-the-art DG approaches face the problem of overfitting on the source domain, thus reducing the generalization abilities, especially if the distribution mismatch between $S$ and $T$ is large (Wang et al., 2021). Regularization and data augmentation techniques as in (Huang et al., 2021; Li et al., 2017; Wang et al., 2020; Xu et al., 2014; Zhou et al., 2020) can increase the generalization abilities but may also increase the difficulty of the learning task, therefore increasing the risk of source-domain overfitting (Wang et al., 2021). The use case considered in this paper assumes the data of non-defective target domain instances to be already available during model training. There are some investigations on the improvement of the DG model's generalization abilities by mixing multiple source domain datasets during training (e.g., (Wang et al., 2020)). However, the specific dataset constellation considered in this paper has not been brought up in research yet.

The highly related Domain Adaptation (DA) approaches assume that unlabeled target domain data are available during training (Kim et al., 2022; Wang et al., 2020; Zhou et al., 2021). As a special case of transfer learning (Pan & Yang, 2010), the idea of DA is to generalize a model which is pre-trained in the source domain, using unlabeled target domain data (Kim et al., 2022). The strong assumption that unbiased target domain data (including defective and non-defective data) are available during training is often not satisfied in practice (Zhou et al., 2021) and does not coincide with the use case considered in this paper: We assume that the target domain data given during training are only from the non-defective class. Classical DA approaches would suffer from the highly biased target domain data which entirely omit defective data. Our target domain data used during training carries additional information in that it includes the class label *non-defective*. However, this additional information is not yet leveraged by traditional DA approaches designed to work with non-labeled target domain data (Kim et al., 2022).

### 2.3. Contrastive Learning

Contrastive Learning (CL) is based on a neural network encoder that outputs a discriminative low-dimensional representation (= embedding or feature vector) of an input sample (Jaiswal et al., 2020). The encoder is trained in a way that related samples (e.g., images of the same class or augmentations of the same image) are aligned in the latent space, while unrelated samples (e.g., images of different classes) are separated (Chen et al., 2020; Jaiswal et al., 2020). A similarity metric is used to measure the distance between two embeddings (Jaiswal et al., 2020).

According to (Thota & Leontidis, 2021) CL has become a key approach for unsupervised learning tasks with unlabeled datasets (Chen et al., 2020; K. He et al., 2019; Jaiswal et al., 2020). Broadly, one sample from the unlabeled dataset is taken as a so-called *anchor* and a strongly augmented version of this sample is considered a *positive* sample. The rest of the samples in the batch are considered *negative* samples, regardless of their actual class (Shen et al., 2022). The encoder is trained in a way that it learns to increase the latent space distance of *anchor* and *negative* samples while decreasing the distance of the strongly related *positive* and *anchor* embeddings. (Jaiswal et al., 2020; Shen et al., 2022) Here, train and test samples are from the same dataset and distribution. A major problem of the contrastive self-supervised architecture is the occurrence of false negatives in the *negative* samples (Thota & Leontidis, 2021).

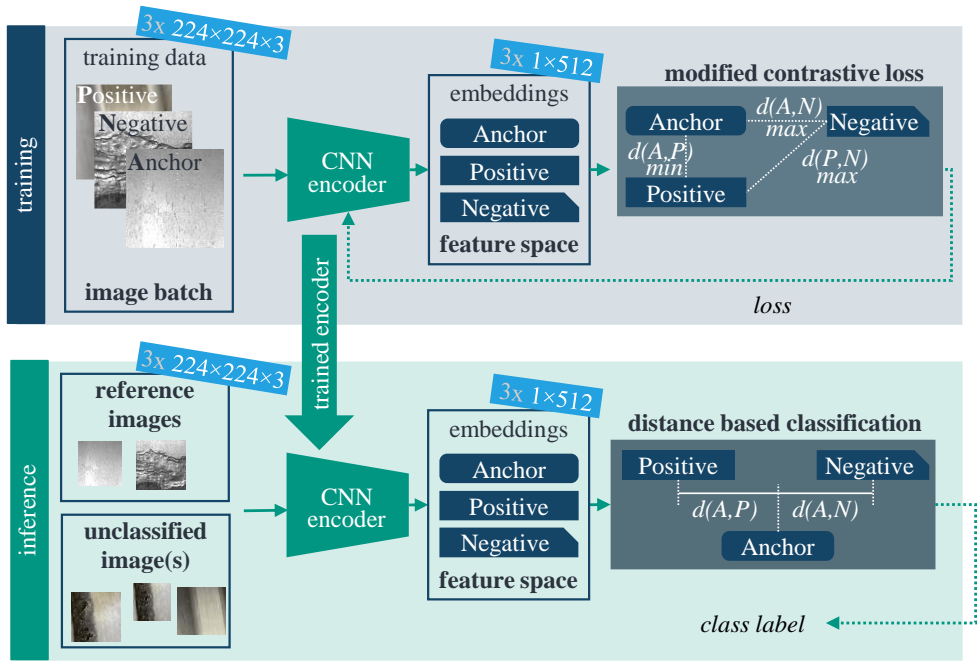

Figure 3: Model architecture of the proposed classifier

Other CL research activities focus on contrastive approaches for DA and DG problems (e.g., (Shen et al., 2022; Thota & Leontidis, 2021; Yang et al., 2022)). A key advantage of CL approaches (concerning the distribution shift from source to target domain) is their robustness against overfitting the source domain samples (Xue et al., 2022). (Shen et al., 2022) use unlabeled target domain and partially labeled source domain data during training. First, an encoder is pretrained using both, unlabeled source, and target data to minimize the distance over augmentations of the same input and maximize the distance over random pairs of inputs (Shen et al., 2022). Afterwards, a projection head is finetuned using the labeled source data (Shen et al., 2022). (Thota & Leontidis, 2021) follow an entirely unsupervised approach, by training only with unlabeled data. However, so far these applications remain largely underexplored in research (Thota & Leontidis, 2021). Furthermore, in the technical domains considered in this paper, different boundary conditions apply: Besides a labeled source-domain dataset, there are samples of the non-defective class of the target domain given in training. This setting is not yet considered in related work and offers unprecedented opportunities:

- There are labeled source domain data from which *anchor* and *negative* samples can be chosen. This eliminates the fundamental problem of false-negative *negatives* during training described by (Thota & Leontidis, 2021).
- Sampling *positives* from the non-defective class of the target domain dataset eliminates the need for data augmentation for the synthetic generation of *positive* samples. Therefore, the model is systematically faced with the domain shift during training since *anchors* and *positives* are chosen from different domains. Using a modified loss function described in the next chapter, we aim to leverage the multi-domain training data to improve the generalization abilities regarding the unseen defective class of the target domain.
- The underlying hypothesis of the decomposability of a defective sample from a technical domain into domain-independent "defect features" and domain-specific "base features" is also stated, used, and proven by (Rombach et al., 2022) in the context of controlled synthetic data generation for out-of-domain defective class samples.

The main contribution of this paper is a modified triplet loss function for classical contrastive learning approaches together with using non-defective data from the target domain. It allows to exploit the previously described opportunities to transfer the classification features across large domain gaps. Chapter 3 describes the general framework, the modified triplet loss function, and the datasets used for the validation of the proposed approach. The model is then analyzed in technical and non-technical scenarios and the results are compared to multiple state-of-the-art benchmarks. The approach shows convincing results and distinctly outperforms the benchmark models. The discussion of the results is followed by a final summary.

## 3. OWN APPROACH

The modified contrastive learning approach presented in this paper addresses classification tasks (e.g., distinguishing defective from non-defective components) in the presence of insufficient target domain training data. Aiming for improved generalization abilities, the model is trained using easily accessible target domain data of the non-defective class, as well as a state-of-the-art source domain dataset containing the same defect in a different domain. However, target domain data of the defective class, showing the feature of interest is not used for training. Our approach is tested and analyzed on different domains and its performance is benchmarked with other state-of-the-art classifiers.

### 3.1. Formalization of the cross-domain generalizing classification task

Following (Kim et al., 2022) there is a source-domain $S$ and a target-domain $T$ given by $S = \{(x_S^k, y_S^k)\}_{k=1}^{K}$ and $T = \{(x_T^m, y_T^m)\}_{m=1}^{M}$, where $x_S$ is a vector of $K$ images and $x_T$ is a vector of $M$ images. $y_S$ and $y_T$ are vectors of corresponding (binary) class labels respectively. The value range of the labels depends on the task and is often given by $y_T^m \in [defect, noDefect]$ in the case of binary tasks in technical domains. Since this paper has a technical background, the following notation is used hereinafter for all domains:

$S_{train} = \{(x_S^k, y_S^k)\}_{k=1}^{K \cdot i - 1} = S_{train,defect} \cup S_{train,noDefect}$
$S_{test} = \{(x_S^k, y_S^k)\}_{k=K \cdot i}^{K} = S_{test,defect} \cup S_{test,noDefect}$
$T_{train} = \{(x_T^m, y_T^m)\}_{m=1}^{M \cdot j - 1} = T_{train,defect} \cup T_{train,noDefect}$
$T_{test} = \{(x_T^m, y_T^m)\}_{m=M \cdot j}^{M} = T_{test,defect} \cup T_{test,noDefect}$

with train-test-split $i, j \in (0,1)$.

The objective is to train a classifier that predicts the label $y_T^m$ of a test image $x_T^m$ from the target domain $T_{test}$. The model is trained using $S_{train,defect}$, $S_{train,noDefect}$, and $T_{train,noDefect}$ and a specific triplet loss function. Since $T_{train,defect}$ is not available during training, the model must generalize well across domains to classify samples from $T_{test}$ correctly.

### 3.2. Datasets

Our research is initially conducted in technical domains and afterward verified in non-technical domains. The technical domain is represented by modified samples of the Severstal steel defect dataset (Severstal, 2020) and a selection of the Ball Screw Drive (BSD) Surface Defect Dataset for Classification (Schlagenhauf, 2021).

The Severstal dataset contains images of strip steel surfaces showing either no defect or defects from six different classes. Simulating a state-of-the-art source domain dataset, we cropped the samples into smaller 224×224 pixel images showing either no defect (21806 pcs) (see Figure 4: Cross-domain Contrastive Learning using modified Triplet Loss function, $S_{train,noDefect}$) or defects of type *patch* (2018 pcs) (see Figure 4: Cross-domain Contrastive Learning using modified Triplet Loss function, $S_{train,defect}$). This defect class was selected since it has the greatest similarity to the pitting defects of the BSD dataset. The BSD dataset on the other hand simulates a target domain dataset. The dataset contains 150×150 pixel RGB images of the surface of ball screw spindles. We use a selection of 1896 images showing no surface defects and little oil contamination as non-defective target domain samples (see Figure 4, $T_{train,noDefect}$). A selection of 5240 images showing pitting defects due to extensive wear forms the defective target domain class. This class is assumed to be not available during training. Even though the two datasets show different objects with different shapes and surface characteristics, the defect feature itself shares the main characteristics in both datasets, making them a suitable pair of source and target domain datasets. The datasets are depicted in Figure 2.

The non-technical domain is represented by an apple leaves dataset and a bean leaves dataset. The apple leaves dataset is a subset of the PlantVillage database that contains 54309 256×256 pixel RGB images of 29 different classes of healthy and diseased plant leaves from 14 different species (Hughes & Salathe, 2015). The apple dataset consists of 1645 healthy images and only 276 diseased images (apple rust) and acts as the source domain dataset. The bean leaves dataset consists of 1296 images with a size of 500×500 pixels. The dataset is divided into one healthy and two diseased classes, of which the angular leaf spot disease class is used for our research due to its greater visual similarities to the disease characteristics of the apple leaves (Makerere AI Lab, 2020). The bean leaves were recorded in nature, while the apple leaves were recorded under laboratory conditions. Consequently, the two datasets differ not only in leaf shape, -color, and -surface texture but also in terms of background and lighting conditions, as depicted in Figure 11. The bean leaves dataset simulates the target domain. All images are scaled to a size of 224×224 pixels and each dataset is split into a train, validation, and test set.

### 3.3. Model Architecture

The model architecture of our binary classifier is depicted in Figure 3. It consists of a trainable CNN encoder that outputs a 512-dimensional embedding vector from any given image of size 224×224. An optional projection head can compute optimized projection vectors from the embeddings. The model is trained using a *non-defective* target domain image (*Positive*), a *defective* source domain image (*Negative*), and a *non-defective* source domain image (*Anchor*). A modified contrastive triplet loss function which quantifies the model's training performance is used to optimize the encoder (and the projection head if applicable). During inference, the model is used to predict class labels of unlabeled *Anchor* images from the target domain (= classification task). The previously trained encoder generates embedding vectors from the *Anchor* image and the two reference images (*Positive* and *Negative*). The *Positive* belongs to the *non-defective* class of source or target domain. The *Negative* is part of the defective *source domain* class since defective target domain samples are not available at the beginning of the inference phase. A distance metric is used to determine the distances $d(A,P)$, $d(A,N)$ between the embeddings/projections of the *Anchor* image and the *Positive* and *Negative* reference images. The class of the Anchor image is then determined based on the shorter of the two distances. To increase robustness against outliers, a single anchor image is classified using many positives and negatives by averaging the individual distances at the end.

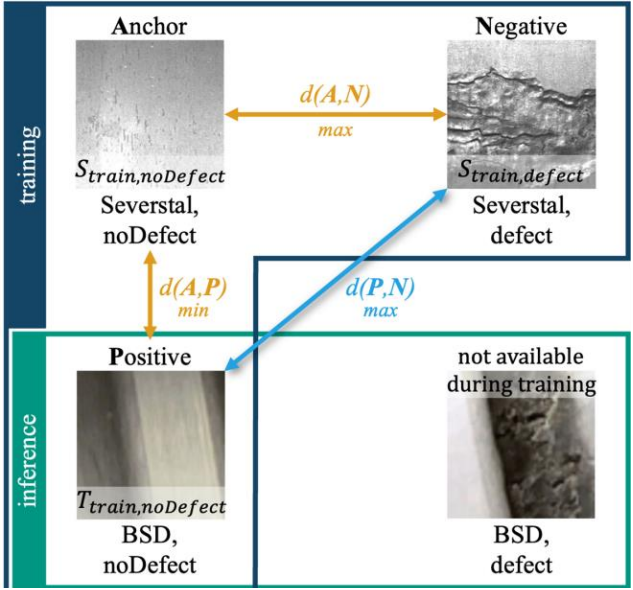

Figure 4: Cross-domain Contrastive Learning using modified Triplet Loss function

### 3.4. Enhancement of the triplet loss function for cross-domain transfer of features

The encoder is trained with a contrastive triplet loss function such that images of the same class lead to related embeddings and images of different classes lead to diverging embeddings - regardless of their domains.
The original triplet loss function is given by

$$L = \max(d(A,P) - d(A,N) + m_1, 0) \quad (1)$$

With distance $d(A,P)$ between **A**nchor and **P**ositive embeddings and distance $d(A,N)$ between **A**nchor and **N**egative embeddings. The margin parameter $m_1$ maintains a minimum difference between the Positive and Negative classes, resulting in self-contained, non-mixed embedding clusters. $d(.)$ can be any similarity metric, however, cosine similarity or Euclidean distance are commonly used. Figure 4 visualizes the loss function by the orange arrows. By minimizing $L$ during training, $d(A,N)$ *is maximized* while $d(A,P)$ *is minimized*. Therefore, embeddings of images of the same class (Severstal noDefect & BSD noDefect) become less distant from each other (despite their different domains) while embeddings of images of different classes (Severstal noDefect & Severstal defect) diverge.

Figure 5: Confusion matrix of the BSD defect classifier that uses our improved triplet loss and is trained on the Severstal strip steel defect dataset and the non-defective class of a BSD defect dataset

In addition to the source domain data, the proposed model leverages the easily accessible target domain samples of the non-defective target domain class $T_{test,noDefect}$ during training, as depicted in Figure 4. Therefore, the *Positives* belong to the target domain, while the *Anchors* belong to the source domain, resulting in two different domains representing one class of non-defectives. By utilizing this setup, we aim for improved domain generalization abilities. However, using the basic contrastive triplet loss in this setup would result in an encoder that learns to disregard domain-related dissimilarities between *Anchor* (source domain) samples and *Positive* (target domain) samples to comply with the distance minimization of $d(A,P)$. This could lead to an oversimplified encoding of the target domain samples.

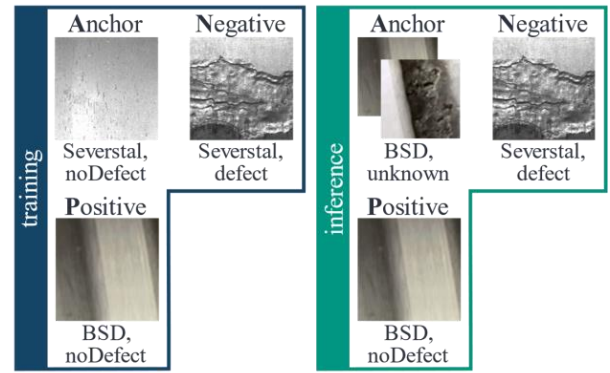

Figure 6: Dataset setup during training and inference of initial test with Severstal strip steel and BSD datasets

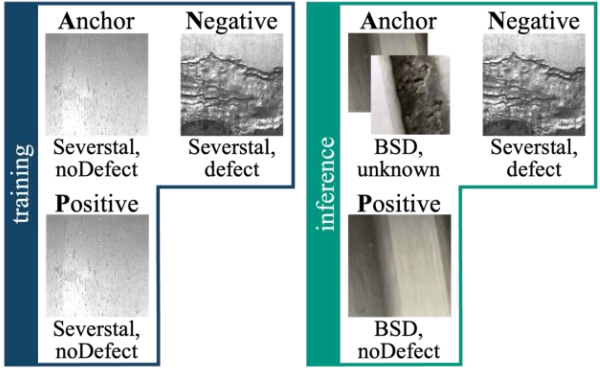

Figure 7: Dataset setup for the benchmark test with basic triplet loss and training on source-domain data only

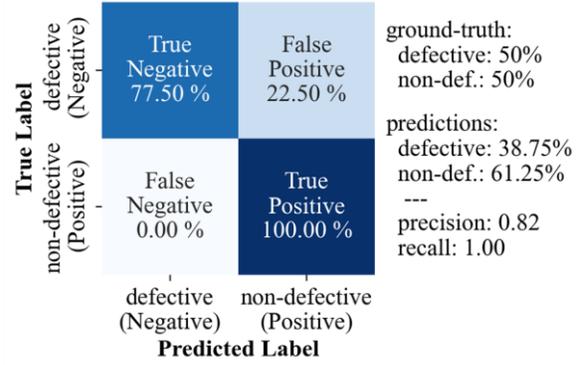

Figure 8: Confusion matrix for the benchmark test of a BSD defect classifier that uses basic triplet loss and is solely trained on the Severstal strip steel defect dataset

Especially if the domain shift between *Anchor* (source domain) and *Positive* (target domain) is large compared to the interclass difference between the *defective* and *non-defective* classes, an oversimplification by the encoder is likely. This can result in a classification bias towards the *non-defective* class since target domain samples are oversimplified as *Positives* while disregarding their actual class, as we will show in Chapter IV.

To address this issue, we extend the basic triplet loss by an additional term that yields Eq. 2. We additionally demand the *maximization* of the distance between the *non-defective* target-domain embeddings (*Positives*) and the *defective* source-domain embeddings (*Negatives*) (cyan arrow in Figure 4), while still demanding a small distance between the two *non-defective* embeddings (*Positives* and *Anchors*). This directs the focus away from domain-specific features towards class-specific features by aiming for distance maximization in the case of different classes (*Positives* & *Negatives*), while distance minimization is aimed for identical classes (*Positives* & *Anchors*). Whether the distance of two samples is maximized or minimized is thus independent of their domains and depends only on their classes:

$$\min L = \min \left( \max(d(A,P) - d(A,N) + m_1, 0) \right. \quad (2)$$
$$\left. + \max(d(A,P) - d(P,N) + m_2, 0) \right)$$

The colors of the two summands match the colors of the arrows in Figure 4. By explicitly maximizing the distance between *Positives* and *Negatives* $d(P,N)$ while demanding a small *Anchor-Positive*-Distance, we aim for a large interclass distance regardless of the domain shift from source to target domains. Our research was conducted using the Euclidean distance metric $d(.)$. Trained using $S_{train,noDefect}$, $S_{train,defect}$ and $T_{train,noDefect}$ the model then classifies unlabeled $T_{test}$ samples with positive and negative reference images, either chosen from $S_{test}$ or $T_{test}$. The influence of the reference image datasets will be further investigated below.

## 4. RESULTS AND DISCUSSION

### 4.1. Experimental results of contrastive learning approaches

#### 4.1.1. Scenario I – Steel defect classification

The proposed approach is analyzed in a typical technical scenario (Schlagenhauf, 2021): The task is to classify BSD spindle images (target domain) into *defectives* and *non-defectives*. The model is trained using *Anchors* from the non-defective and *Negatives* from the defective class of the state-of-the-art Severstal steel defect dataset respectively (= source domain). The *Positives* belong to the non-defective class of the BSD dataset (= target domain). The dataset setup during training and inference is depicted in Figure 9. The model is trained without defective target domain samples at all. Using 1000 *Positives*, *Negatives*, and *Anchors* each, the model is trained for 150 epochs with a batch size of 60 and the modified contrastive triplet loss function (Eq. 2). The margin parameters $m_1$ and $m_2$ are set to 0.2. In pre studies with 10 iterations per experiment on the BSD and Severstal datasets, the authors explored a small variance in the results. Once training is completed, the non-defective Severstal *Anchor* images are swapped for unlabeled target domain (BSD) images in the inference phase. The trained encoder generates low-dimensional embeddings from 200 defective Severstal *Negatives* and 200 non-defective BSD *Positives* (= reference images), as well as from 80 equally distributed BSD test images (*Anchors*). Each of the 40 defective and 40 non-defective BSD test samples is classified based on its Euclidean distance to the 200 positive and 200 negative embeddings.

According to the confusion matrix (Figure 5), 100 % of the non-defective BSD samples and 95 % of the defective BSD samples are correctly classified. The test results provide evidence that the model can learn the characteristic features of the defect from the Severstal strip steel dataset (source-domain) and can successfully perform the domain shift towards the BSD samples (target-domain).

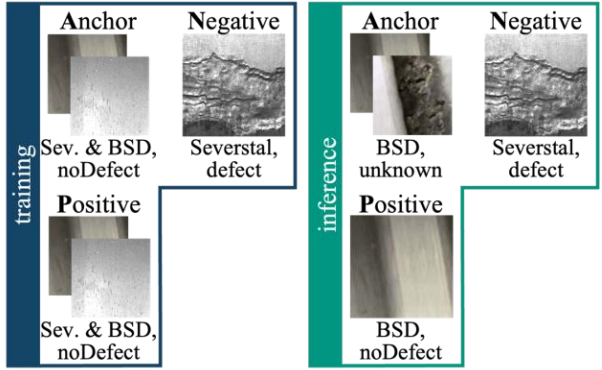

Figure 9: Modified training setup for the second benchmark test using basic triplet loss and source & target domains for training

In the following, we present two benchmark tests that demonstrate the performance improvements of our proposed approach. The first benchmark model is trained using only Severstal strip steel data with the basic, unmodified triplet loss function (Eq. 1). Figure 7 depicts the modified training setup. The inference setup remains untouched. The confusion matrix shown in Figure 8 testifies the inferior domain generalization abilities of the benchmark model compared to our proposed model: While all non-defective images are still correctly classified, more than four times as many defective images are misclassified as non-defective, leading to a False Positive rate of 22.5 %. Trained exclusively with source domain data, the benchmark model shows inferior domain generalization capabilities compared to the proposed model. This results in a classification bias towards the non-defective class, leading to the increased False Positive rate.

Further studies show that the inference performance of the benchmark model is even decreased if the *Positives* are no longer generated from *target* domain samples but from *source* domain samples. In this case, the False Positive rate further increases from 22.5 % to 60 %.

Compared to the proposed approach, the previous test (first benchmark) was deprived of information, since the healthy target domain class was not given during training. Therefore, the following benchmark experiment is trained with the same data as our proposed approach: The *Anchors* and *Positives* are composed of the non-defective target domain- and source domain classes. However, the benchmark model does not use the modified loss function. The inference setup remains untouched again.

The previously observed classification bias towards the non-defective class is further increased by training with the basic triplet loss. According to Figure 10 the False Positive rate increases from 22.5 % (previous benchmark test) to 75 %. Even though the model's recall of 0.97 is good, its low precision makes the model unusable for this classification task.

Figure 10: Confusion matrix of the second benchmark test using basic triplet loss and training on the Severstal strip steel defect dataset and the non-defective class of a BSD defect dataset

The three experiments show that our proposed approach can significantly outperform the classification performance of the two benchmark models by systematic training with additional data from the target domain and the modified triplet loss (Eq. 2).

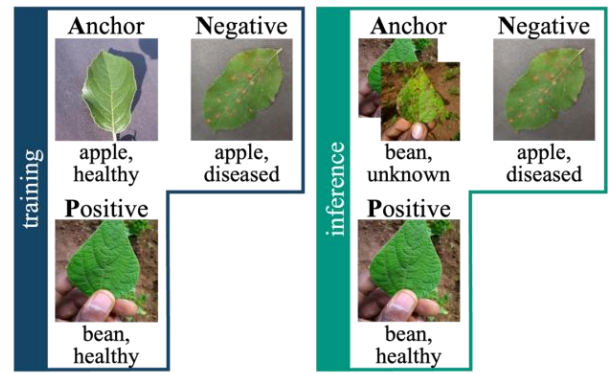

Figure 11: Dataset setup during training and inference of initial test with apple and bean leaves datasets.

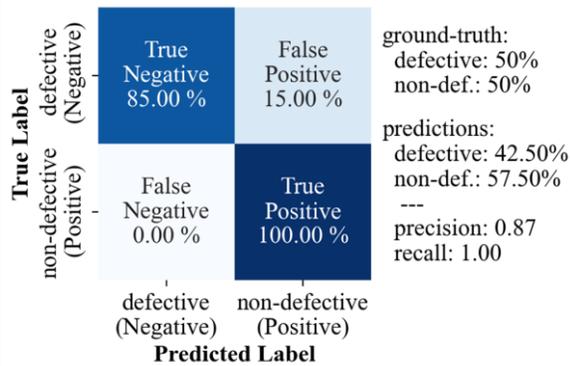

Figure 12: Confusion matrix of the bean leaves classifier that uses our improved triplet loss and is trained on an apple leaves dataset and the healthy class of a bean leaves dataset.

### 4.1.2. Scenario II - Leafe disease classification

In the following, the generalization abilities of the proposed approach are further investigated in non-technical domains using two plant leaf datasets: The model is trained with just 275 sample images of healthy apple leaves as *Anchors*, 275 samples of diseased apple leaves as *Negatives* and 275 samples of the healthy (non-defective) class of a bean leaves dataset as *Positives* (Hughes & Salathe, 2015), (Makerere AI Lab, 2020). There is a significant difference in the context of the two domains, whereas the diseased spot features (brown spots) appear similar in both domains, as depicted in Figure 11.

The model is trained for 200 epochs with a batch size of 60 and the modified triplet loss function given by (Eq. 2). Despite the small training dataset size, no overfitting was observed, which is also consistent with (Xue et al., 2022). The authors attribute this to the random selection of triplets

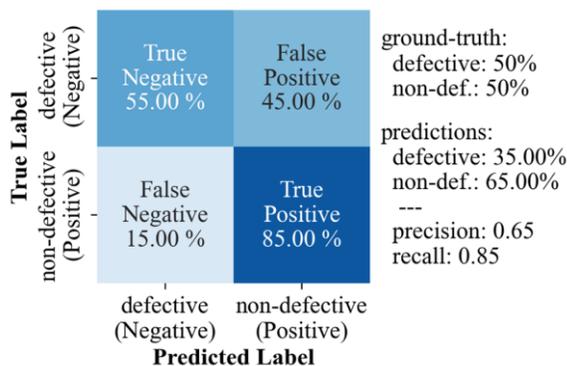

Figure 13: Confusion matrix of the second benchmark test using basic triplet loss and training on an apple leaves dataset and the healthy class of a bean leaves dataset

from *Anchors*, *Positives*, and *Negatives*, maximizing chances that no triplet occurs more than once in the same constellation during training.

The trained encoder generates low-dimensional embeddings from 80 diseased apple leaves (*Negatives*) and 80 healthy bean leaves (*Positives*) (= reference images), as well as from 80 equally distributed bean leaf test images (*Anchors*) (see Figure 11). Each of the 40 diseased and 40 healthy bean test samples is classified based on its Euclidean distance to the 80 positive and 80 negative embeddings. According to the confusion matrix (Figure 12), 100 % of the healthy bean leaves and 85 % of the diseased bean leaves are correctly classified. The test results provide evidence that the model can learn the characteristic features of the disease from the apple leaves dataset (source-domain) and can successfully perform the domain shift towards the target domain (bean leaves). Furthermore, the domain generalization abilities are improved by the provided healthy target domain class samples as the following benchmark test testifies:

The benchmark model is trained using only apple leaves data with the basic triplet loss function (Eq. 1). Figure 14 depicts the modified dataset setup during training and the unchanged inference setup.

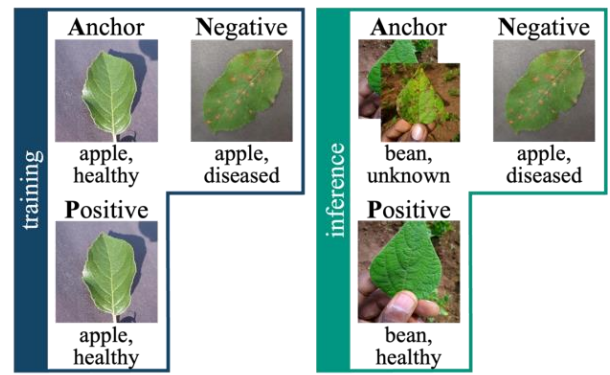

Figure 14: Modified training setup for the benchmark test with basic triplet loss and training on source-domain data only

The confusion matrix shown in Figure 15 testifies the poor domain generalization abilities of the benchmark test. Only 55 % of the healthy – and 62.5 % of the diseased bean leaves test dataset samples are correctly classified. The False Positive Rate is more than doubled compared to our proposed approach. Almost every second healthy ("non-defective") test sample is falsely classified as diseased ("defective"), whilst our proposed approach correctly classifies all healthy test images.

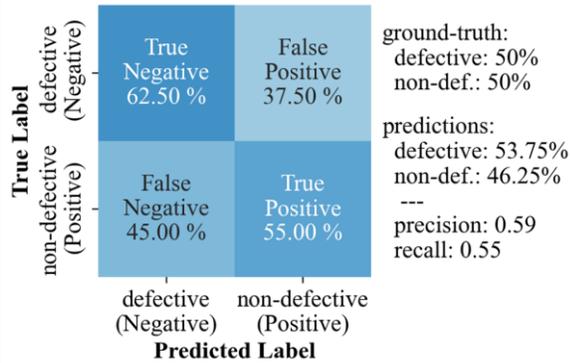

Figure 15: Confusion matrix of the bean leaves classifier that uses basic triplet loss and is solely trained on an apple leaves dataset.

Accordingly, the benchmark model trained only with source domain data is not able to generalize to the target domain. Again, further studies show that the inference performance of the benchmark model is decreased if the *Positives* are no longer generated from *target* domain samples but *source* domain samples. In this case, the False Positive rate further increases from 37.5 % to 92.5 %.

Again, the benchmark model was deprived of information since the healthy target domain class is not given during training. Therefore, the next benchmark experiment uses the same model and basic loss function as the previous benchmark, however, now trained on the same database as the initial plant leaves experiment, thus combining source and target domain data. The dataset setup is depicted in Figure 16. Even though the recall improves from 0.55 to 0.85 due to the additional training data, the model's ability to classify diseased bean leaf samples decreases as posed by the confusion matrix (Figure 13). With a False Positive rate of 45 %, the model is again highly biased towards the positive (healthy) class. The three experiments show that the findings from the technical domain also apply to the non-technical domain.

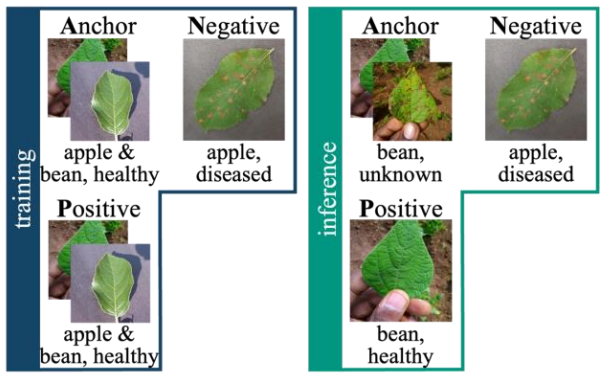

Figure 16: Modified training setup for the second benchmark test using basic triplet loss and source & target domains during training.

**4.2. Basic object detection benchmark**

A very different yet common image classification approach uses an object detector that classifies a sample as defective if at least one defect is detected. The main drawback of state-of-the-art approaches often is the insufficient generalization ability if source and target domains differ too much. To analyze the model's performance in the present case, we confronted a state-of-the-art region-based Convolutional Neural Network (R-CNN) object detection model with the domain transfer from BSD to Severstal data. Trained in the BSD domain, the model provides good classification results within this domain. However, the model is not able to generalize enough to manage the domain shift from BSD to Severstal data: None of the defects on the Severstal samples are detected. The findings indicate that the generalization abilities of the state-of-the-art object detection model are insufficient for the use case considered here.

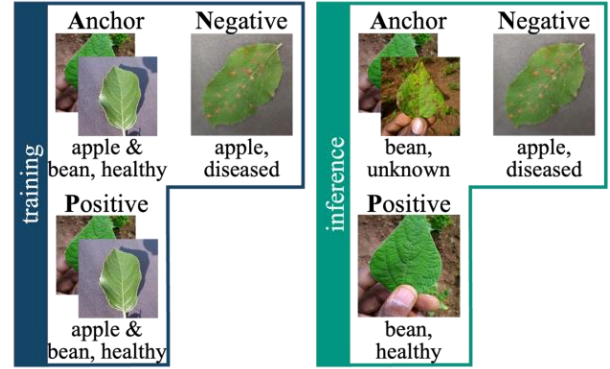

Figure 17: Modified training setup for the second benchmark test using basic triplet loss and source & target domains during training

**4.3. Discussion: Domain generalization without target domain data**

Table 1 summarizes the results of the presented experiments. The proposed contrastive learning model (experiments 1, 5) was developed for applications that require the model to be trained without a complete target domain dataset. Learning from a source domain dataset and the early available target domain data, the model must generalize across large domain disparities. Experiments 2 and 6 prove that this model is not able to generalize sufficiently when trained on source-domain data only (without target domain data).

| # | EXPERIMENT | TRUE POSITIVE | FALSE NEGATIVE | TRUE NEGATIVE | FALSE POSITIVE | CONTRAST. LOSS | TRAINING DATA | | |
|---|---|---|---|---|---|---|---|---|---|
| 1 | Steel Defects, Our Model | 100.0 % | 0.0 % | 95.0 % | 5.0 % | Eq. 2 | A: $S_{train,noDef}$, | P: $T_{train,noDef}$, | N: $S_{train,defect}$ |
| 2 | Steel Defects, Benchmark 1 | 100.0 % | 0.0 % | 77.5 % | 22.5 % | Eq. 1 | A: $S_{train,noDef}$, | P: $S_{train,noDef}$, | N: $S_{train,defect}$ |
| 3 | Steel Defects, Benchmark 2 | 97.5 % | 2.5 % | 25.0 % | 75.0 % | Eq. 1 | A: $S, T_{train,noDef}$, | P: $S, T_{train,noDef}$, | N: $S_{train,defect}$ |
| 4 | Steel Defects, Obj. Detection | 100.0 % | 0.0 % | 0.0 % | 100.0 % | - | Train: $S_{train}$, | Test: $T_{train}$ | |
| 5 | Leaves, Our Model | 100.0 % | 0.0 % | 85.0 % | 15.0 % | Eq. 2 | A: $S_{train,noDef}$, | P: $T_{train,noDef}$, | N: $S_{train,defect}$ |
| 6 | Leaves, Benchmark 1 | 55.0 % | 45.0 % | 62.5 % | 37.5 % | Eq. 1 | A: $S_{train,noDef}$, | P: $S_{train,noDef}$, | N: $S_{train,defect}$ |
| 7 | Leaves, Benchmark 2 | 85.0 % | 15.0 % | 55.0 % | 45.0 % | Eq. 1 | A: $S, T_{train,noDef}$, | P: $S, T_{train,noDef}$, | N: $S_{train,defect}$ |

Table 1: Experimental results

### 4.4. Discussion: The modified loss function leverages additional target domain training data to improve domain generalization

Additionally providing non-defective target domain class data ($T_{train,noDef}$) during the training of the base model (basic triplet loss, Eq. 1) strongly increases the false positive rate as demonstrated by experiments 3 and 7. This classification bias limits the model's usability. To leverage the additional target domain data and improve the generalization capability, the proposed model with its modified loss function (see Eq. 2) must be used for training, as testified by experiments 1 and 4. The modified loss function (Figure 4) allows the model to:

- use the source-domain dataset with its defective ($N$) and non-defective ($A$) classes to systematically learn the features that characterize the defect (by maximizing $d(A, N)$).
- use the non-defective source-domain class ($A$) and non-defective target-domain class ($P$) to suppress the (extraneous) context, that differs between source and target domains (by minimizing $d(A, P)$).
- prevent the target domain classification bias towards the non-defective class ($P$) as a result of the one-sided training with non-defective target domain samples ($P$) (by maximizing $d(P, N)$).

The modified triplet loss facilitates systematic learning of the domain-independent features of defects, regardless of the domain-specific context.

### 4.5. Discussion: Data efficiency and robustness against overfitting

With 275 samples per dataset, the training datasets of the non-technical domain are small. Even without a pre-trained network, our model is still able to learn the classification task sufficiently well (see experiment 5). Furthermore, no overfitting was observed even with such limited training data. This is attributed to the specific characteristics of the contrastive learning approach: The random selection of triplets from *Anchors*, *Positives*, and *Negatives* mostly ensures that no triplet occurs more than once in the same constellation during training. This leads to a large variety of training triplets, where each image can be used multiple times in different triplet constellations.

### 4.6. Discussion: Gauge of the Domain Gap

In the here presented approach, the authors did experiments with two types of domains, where the classes within the domains are similar (leafes, metallic surfaces). The selection of the suitable source- and target-domain datasets was done based on domain knowledge. There has been no other similarity measurement with respect to the similarity of the domains. Though, this may be one main research direction. It is especially interesting, how similar two domains must be such that the here presented approach works. Here two interesting research directions emerge. First, it would be very interesting what similarity means with respect to different classes. This may help to understand similarity in a semantic way. The authors suggest choosing N different datasets with decreasing similarity (assessed by the authors) and to do the above experiments. As a result, a NxN similarity matrix is received. The second very interesting question which can also potentially be answered with the same procedure is the question on how similar two datasets must be such that the proposed approach works.

### 4.7. Discussion: Other promising research directions

Well knowing that some entitled questions remain open the authors wanted to show a promising avenue in the field of domain generalization. Additional to the before mentioned points, the proposed approach only addresses binary classification tasks so far. In the next step, the approach will be extended to multi-class problems. Another research topic will be the analysis of different loss functions that follow the same basic idea. Furthermore, instead of classification problems, the proposed approach will be integrated into object-detection problems.

## 5. CONCLUSION

We proposed a modified contrastive learning approach with an extended triplet loss function. The approach targets binary classification tasks where not all target domain classes are available during training. Therefore, an additional state-of-the-art source domain dataset that contains all classes and shows the same defect/ classification feature but originates from a (highly) different domain is involved in training. By the combination of both, the source and the (partial) target domain datasets, our model systematically learns the relevant features for the classification and masters the domain shift from source to target domain. This was analyzed in two different use cases. The experimental results demonstrate that our proposed approach leverages the partial target domain data which is already available during training and outperforms a state-of-the-art object detection-based classifier and contrastive learning approaches. Compared to zero-shot methods, which only train with source-domain data, and one-shot/few-shot methods, which usually require target domain samples of all classes (especially of the "defective class"), our model for the first time can make good use of the limited target domain data in form of just one class. Therefore, it can be trained and used at an early stage for binary classification when other models either cannot be trained yet or cannot achieve sufficient classification results with the limited data available.

## 6. REFERENCES

Ben-David, S., Blitzer, J., Crammer, K., Kulesza, A., Pereira, F., & Vaughan, J. W. (2010). A theory of learning from different domains. *Machine Learning*, *79*(1–2), 151–175. https://doi.org/10.1007/s10994-009-5152-4

Blanchard, G., Lee, G., & Scott, C. (2011). *Generalizing from Several Related Classification Tasks to a New Unlabeled Sample*. https://papers.nips.cc/paper/2011/file/b571ecea16a9824023ee1af16897a582-Paper.pdf

Chen, T., Kornblith, S., Norouzi, M., & Hinton, G. (2020). *A Simple Framework for Contrastive Learning of Visual Representations*.

Chohan, M., Khan, A., Chohan, R., Katpar, S. H., & Mahar, M. S. (2020). Plant Disease Detection using Deep Learning. *International Journal of Recent Technology and Engineering (IJRTE)*, *9*(1), 909–914. https://doi.org/10.35940/ijrte.A2139.059120

Hamadache, M., Jung, J. H., Park, J., & Youn, B. D. (2019). A comprehensive review of artificial intelligence-based approaches for rolling element bearing PHM: shallow and deep learning. *JMST Advances*, *1*(1–2), 125–151. https://doi.org/10.1007/s42791-019-0016-y

He, K., Fan, H., Wu, Y., Xie, S., & Girshick, R. (2019). *Momentum Contrast for Unsupervised Visual Representation Learning*.

He, Y., Shen, Z., & Cui, P. (2019). *Towards Non-I.I.D. Image Classification: A Dataset and Baselines*.

Huang, J., Guan, D., Xiao, A., & Lu, S. (2021). *FSDR: Frequency Space Domain Randomization for Domain Generalization*.

Hughes, David. P., & Salathe, M. (2015). *An open access repository of images on plant health to enable the development of mobile disease diagnostics*. https://data.mendeley.com/datasets/tywbtsjrjv/1

Jaiswal, A., Babu, A. R., Zadeh, M. Z., Banerjee, D., & Makedon, F. (2020). *A Survey on Contrastive Self-supervised Learning*. http://arxiv.org/abs/2011.00362

Kaiser, J.-P., Mitschke, N., Stricker, N., Heizmann, M., & Lanza, G. (2021). Konzept einer automatisierten und modularen Befundungsstation in der wandlungsfähigen Produktion. *Zeitschrift Für Wirtschaftlichen Fabrikbetrieb*, *116*(5), 313–317. https://doi.org/10.1515/zwf-2021-0070

Kim, D., Wang, K., Sclaroff, S., & Saenko, K. (2022). *A Broad Study of Pre-training for Domain Generalization and Adaptation*.

Li, D., Yang, Y., Song, Y.-Z., & Hospedales, T. M. (2017). *Deeper, Broader and Artier Domain Generalization*.

Makerere AI Lab. (2020). *Bean disease dataset*. https://github.com/AI-Lab-Makerere/ibean/

Maqsood, M., Nazir, F., Khan, U., Aadil, F., Jamal, H., Mehmood, I., & Song, O. (2019). Transfer Learning Assisted Classification and Detection of Alzheimer's Disease Stages Using 3D MRI Scans. *Sensors*, *19*(11), 2645. https://doi.org/10.3390/s19112645

Pan, S. J., & Yang, Q. (2010). A Survey on Transfer Learning. *IEEE Transactions on Knowledge and Data Engineering*, *22*(10), 1345–1359. https://doi.org/10.1109/TKDE.2009.191

Rombach, K., Michau, Dr. G., & Fink, Prof. Dr. O. (2022). *Controlled Generation of Unseen Faults for Partial and Open-Partial Domain Adaptation*.

Schlagenhauf, T. (2021). *Ball Screw Drive Surface Defect Dataset for Classification* (K. I. für Technologie, Ed.). Karlsruher Institut für Technologie (KIT) wbk Institute of Production Science. https://doi.org/10.5445/IR/1000133819

Schlagenhauf, T., Scheurenbrand, T., Hofmann, D., & Krasnikow, O. (2022). *Analysis of the Visually Detectable Wear Progress on Ball Screws*.

Severstal. (2020, December 8). *Severstal: Steel Defect Detection* (Severstal, Ed.). https://www.kaggle.com/c/severstal-steel-defect-detection/data

Shen, K., Jones, R., Kumar, A., Xie, S. M., HaoChen, J. Z., Ma, T., & Liang, P. (2022). *Connect, Not Collapse: Explaining Contrastive Learning for Unsupervised Domain Adaptation*. http://arxiv.org/abs/2204.00570

Siddiqui, M. W. (2015). *Postharvest Biology and Technology of Horticultural Crops*. Apple Academic Press.

Thota, M., & Leontidis, G. (2021). *Contrastive Domain Adaptation*.

Torralba, A., & Efros, A. A. (2011). Unbiased Look at Dataset Bias. *CVPR 2011*, 1521–1528. https://doi.org/10.1109/CVPR.2011.5995347

Wang, Y., Li, H., Chau, L., & Kot, A. C. (2021). Embracing the Dark Knowledge: Domain Generalization Using Regularized Knowledge Distillation. *Proceedings of the 29th ACM International Conference on Multimedia*, 2595–2604. https://doi.org/10.1145/3474085.3475434

Wang, Y., Li, H., & Kot, A. C. (2020). *Heterogeneous Domain Generalization via Domain Mixup*. https://doi.org/10.1109/ICASSP40776.2020.9053273

Xu, Z., Li, W., Niu, L., & Xu, D. (2014). *Exploiting Low-Rank Structure from Latent Domains for Domain Generalization* (pp. 628–643). https://doi.org/10.1007/978-3-319-10578-9_41

Xue, Y., Whitecross, K., & Mirzasoleiman, B. (2022). *Investigating Why Contrastive Learning Benefits Robustness Against Label Noise*.

Yang, C., Cheung, Y. M., Ding, J., Tan, K. C., Xue, B., & Zhang, M. (2022). Contrastive Learning Assisted-Alignment for Partial Domain Adaptation. *IEEE Transactions on Neural Networks and Learning Systems*. https://doi.org/10.1109/TNNLS.2022.3145034

Zhang, X., Cui, P., Xu, R., Zhou, L., He, Y., & Shen, Z. (2021). *Deep Stable Learning for Out-Of-Distribution Generalization*.

Zhang, X., Xu, Z., Xu, R., Liu, J., Cui, P., Wan, W., Sun, C., & Li, C. (2022). *Towards Domain Generalization in Object Detection*.

Zhou, K., Liu, Z., Qiao, Y., Xiang, T., & Loy, C. C. (2021). *Domain Generalization: A Survey*.

Zhou, K., Yang, Y., Hospedales, T., & Xiang, T. (2020). Deep Domain-Adversarial Image Generation for Domain Generalisation. *Proceedings of the AAAI Conference on Artificial Intelligence*, *34*(07), 13025–13032. https://doi.org/10.1609/aaai.v34i07.7003